\documentclass[11pt]{article}
\usepackage[final]{acl}

\usepackage[T2A,T1]{fontenc}
\usepackage[utf8]{inputenc}
\usepackage[russian,english]{babel}
\usepackage{times}
\DeclareFontFamilySubstitution{T2A}{ptm}{cmr}

\usepackage{microtype}

\usepackage{graphicx}
\usepackage{comment}

\usepackage{booktabs}

\usepackage{latexsym}
\usepackage{amsmath}

\usepackage{menukeys}

\usepackage{xspace}
\usepackage{enumitem}
\usepackage{adjustbox}

\usepackage[capitalize,noabbrev]{cleveref}

\ifdefined\tokenmacrosloaded \fi
\let\tokenmacrosloaded\relax

\IfFileExists{latexml.sty}{\usepackage{latexml}}{\newif\iflatexml\latexmlfalse}
\usepackage{xcolor}    % no option list: never clashes with an earlier load

\colorlet{pretokbg}{white}
\colorlet{pretokbr}{black!45}
\colorlet{pretoktx}{black!75}

\iflatexml
\colorlet{tokbg}{black!12}
\colorlet{pretokbghtml}{black!4}   % white would vanish if the CSS is dropped

\lxRequireResource[type=text/css,content={
.tok-key, .tok-key *, .pretok-key, .pretok-key * {
  font-family: ui-monospace, SFMono-Regular, Menlo, Consolas, monospace !important;
}
.tok-key, .pretok-key {
  font-size: 0.9em;
  border-radius: 4px;
  padding: 0.08em 0.32em;
  white-space: pre;
}
.tok-key {
  background-color: #e0e0e0;
  border: 1px solid #b5b5b5;
  border-bottom: 2px solid #9a9a9a;
  color: #111111;
}
.pretok-key {
  background-color: #f5f5f5;
  border: 1px dashed #9a9a9a;
  color: #444444;
}
}]{}

\makeatletter
\newcommand{\tk@box}[3]{% #1=bg color #2=css class #3=content
  \colorbox{#1}{\lxAddClass{#2}\ttfamily#3}%
}
\newcommand{\tk@list}[3]{% #1=bg color #2=css class #3=comma list
  \begingroup
  \def\tk@sep{}%
  \@for\tk@item:=#3\do{%
    \tk@sep
    \tk@box{#1}{#2}{\tk@item}%
    \def\tk@sep{\,}%
  }%
  \endgroup
}
\newcommand{\tokens}[2][]{\tk@list{tokbg}{tok-key}{#2}}
\newcommand{\pretokens}[1]{\tk@list{pretokbghtml}{pretok-key}{#1}}
\makeatother

\else
\usepackage{menukeys}
\renewmenumacro{\keys}[,]{roundedkeys}
\changemenuelement{roundedkeys}{sep}{\hspace{0.1em}}
\tikzset{tw@roundedkeys@base/.append style={font=\ttfamily}}
\newcommand{\tokens}[2][]{%
  \begingroup
  \if\relax\detokenize{#1}\relax\else
    \changemenucolortheme{roundedkeys}{#1}%
  \fi
  \keys{#2}%
  \endgroup
}

\copymenustyle{pretokkeys}{roundedkeys}
\changemenuelement{pretokkeys}{sep}{\hspace{0.3em}}
\tikzset{tw@pretokkeys@base/.append style={font=\ttfamily,dashed}}
\newmenucolortheme{pretok}{named}{pretokbg}{pretokbr}{pretoktx}
\changemenucolortheme{pretokkeys}{pretok}
\newmenumacro{\pretokens}[,]{pretokkeys}
\fi

\newcommand{\tsp}{\textvisiblespace}

\colorlet{mkcol}{RoyalBlue!75!black}
\newcommand{\mkglyph}{\textbrokenbar}
\newcommand{\mk}{\textcolor{mkcol}{\texttt{\mkglyph}}}

\newcommand{\clonehundredk}{\texttt{cl100k}\xspace}

\newcommand{\dx}{\ensuremath{\Delta}\texttt{x}}

\colorlet{bndcol}{OliveGreen!75!black}
\makeatletter
\newcommand{\bnd@key}[1]{\@ifundefined{bnd@#1}{??#1??}{\@nameuse{bnd@#1}}}
\@namedef{bnd@w}{w}
\@namedef{bnd@wp}{w,p}
\@namedef{bnd@wpd}{w,p,d}
\@namedef{bnd@wcaps}{w,\textuparrow}
\@namedef{bnd@wpcaps}{w,p,\textuparrow}
\@namedef{bnd@wpdcaps}{w,p,d,\textuparrow}
\@namedef{bnd@wpdcapsin}{w,p,d,\textuparrow\textsubscript{in}}
\newcommand{\bnd}[1]{\textcolor{bndcol}{{boundary[\bnd@key{#1}]}}}
\newcommand{\bnds}[1]{\textcolor{bndcol}{{[\bnd@key{#1}]}}}
\newcommand{\plainscheme}{\textcolor{bndcol}{plain}}
\makeatother

\colorlet{ccol}{BurntOrange!85!black}
\newcommand{\shf}{\textcolor{ccol}{\texttt{\textuparrow}}}
\newcommand{\cps}{\textcolor{ccol}{\ensuremath{\Uparrow}}}

\title{\pretokens{\shf\mk explicit\mk,\shf\mk boundary\mk,\shf\mk markers\mk}\\ \pretokens{\mk for\mk,\shf\mk subword\mk,\shf\mk vocabularies\mk}}
\author{Sander Land \\
  Neko Health   \\
  \texttt{sander.land@nekohealth.com} 
  \And
  Clara Meister \\
  EPFL \\
  \texttt{clara.meister@epfl.ch} \\
}

\begin{document}
\maketitle

\begin{abstract}
Subword tokenizers represent many common words twice in space-using writing systems, once
with a leading space and once without. The two entries have separate
embeddings, so occurrences of one word are divided across rows that are trained
independently, and the two forms need not even segment the string the same way:
\tokens{\tsp{}together} may be a single entry while the same word without a preceding
space is tokenized as \tokens{to,gether}. Capitalization divides a word further, into as
many as six forms.
We introduce an alternative to standard whitespace conventions using an explicit word boundary marker,
which prevents such duplication.
Words are delimited by the boundary markers, and spaces between words are represented as
pairs of such markers. Two shift codes do the same for title case and upper case.
Switching to this convention mitigates the duplicate-entry issue,
but does not improve compression: for both vocabulary-learning algorithms,
the best marker scheme stays within one percent of the baseline in characters per token,
averaged across six languages.
It does result in better language modeling performance for some languages. 
In English and Russian, language models trained with every marker scheme tested have
lower bits per byte than the baseline, 
suggesting that duplication carries a cost that compression does not capture.
\\
 \href{https://github.com/sanderland/script_tok/}{
 \raisebox{-0.15\height}{\includegraphics[width=0.3cm]{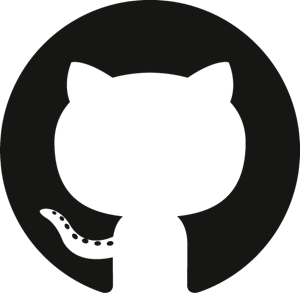}}
 \texttt{\small\,github.com/sanderland/script\_tok}
 }
\end{abstract}

\section{Introduction}
\label{sec:intro}

Most subword tokenizers store two entries for a given word: one whose string begins with a space, and one whose string does not. Which of the two is used depends on the character that precedes the word in the text. 
In \clonehundredk \citep{tiktoken}, ``to be or not to be'' is tokenized as \tokens{to,\tsp{}be,\tsp{}or,\tsp{}not,\tsp{}to,\tsp{}be}, using two different tokens for `to'. 
Storing both is what allows a space to be encoded together with the word that follows it, rather than as a
token of its own.
%CR: \footnote{WordPiece inverts the convention, marking word-internal pieces with a continuationprefix instead of word-initial ones with a space, and duplicates entries just the same.} 
Capitalization produces duplication of the same kind: \tokens{The},
\tokens{the} and \tokens{THE} are typically separate (and therefore unrelated) entries in a vocabulary. Combined, a
word can appear in as many as six forms.\footnote{Among the 10{,}000 most frequent English words in
\citet{speer2022wordfreq}, the 3{,}784 that \clonehundredk stores whole appear in the vocabulary (on average) in 4.2 forms each;
771 appear in all six.}

Duplication splits the training signal for one word. Each form has its own embedding row,
estimated from its own occurrences, so the rarer form of a common word can receive as few
updates as a genuinely rare string.
The two forms may also be segmented completely
differently and share no pieces at all: 5{,}178 words in \clonehundredk are a single token
in one form and split into two pieces (89\%) or more (11\%) in the other. The same word
therefore reaches the model as two unrelated sequences of tokens, depending on the
character that precedes it.
%The distinction also affects prediction: the same word is predicted through different token sequences depending on the character that precedes it.

Our proposed method targets both sources of duplication by changing how word boundaries and
capitalization are represented, before any vocabulary is learned. Boundary markers and case
codes are added as atomic tokens and emitted during pretokenization, replacing the usual
rule that attaches a space to the word after it. The transformation is invertible, so no
information about the original text is lost, and the vocabulary-learning algorithm is
unchanged, so the scheme applies to BPE, Unigram and other tokenization algorithms alike.

We propose and evaluate a range of word boundary and capitalization marking schemes on six languages and two tokenization algorithms,
measuring compression and morphological alignment, and we train language models in English,
Korean and Russian to measure bits per byte. We also train tokenizers on English and Korean
text together, to measure how the schemes change the division of one vocabulary between two scripts.
%morphological alignment and downstream language modeling quality.

\section{Related work}
\label{sec:related}

Word duplication in tokenizers is a known problem.
\citet{chai2024tokenizationfallingshortsubword} report that capitalization variants receive
unrelated token representations, and \citet{ayoobi2026saythistokenizerbetrays} probe models on
token-level rewriting and trace part of the failure rate to tokenizer artifacts, naming
space-duplicated vocabulary entries among them.
\citet{oh-schuler-2024-leading} identify another problem with leading-whitespace tokens:
they confound the calculation of word probabilities, which they address by reallocating
whitespace probability during decoding.

Several solutions have been proposed, but have seen little adoption in modern language models.
\citet{gow-smith-etal-2022-improving} attribute the
inconsistent segmentation of equivalent strings to tokens being allowed to include spaces, and
treat every space as its own token in BPE and Unigram, reporting gains on tasks involving
complex words.
TokenMonster \citep{tokenmonster} uses capitalization codes and forward-delete tokens
to share lexical material across case and spacing variants.
Similarly, \citet{semenov2025inca} use inline markers to separate casing information
from lexical content.
\citet{wilken2019novelapplicationsfactoredneural} address space and case at the NMT output layer, predicting a lowercased subword and a casing class as separate factors, rather than modifying the tokenizer itself.

\citet{wordpiece_boundary} mark whether spaces occur to the left and right of WordPiece units, allowing the original spacing to be reconstructed. This is the closest prior representation to ours, but their markers encode the presence of literal whitespace, so the same whole word can have different forms depending on its surrounding spaces.
Our markers instead encode word-span boundaries unconditionally, giving whole words the same form across surrounding contexts.
Our focus is on eliminating this whitespace-conditioned whole-word duplication, and on measuring its effect while keeping the vocabulary-learning algorithm unchanged.

\section{Methods} 
\label{sec:method}

\paragraph{Definitions.} Our proposed marking method replaces the pretokenizer's space-attachment rule rather
than preceding it, but two levels of segmentation are still involved and we name them
separately. A \emph{span} is a contiguous region of the input selected for marking. A
\emph{pretoken} is a maximal region produced alongside marking, and is the unit over which
the vocabulary is learned. We call a member of that vocabulary an \emph{entry}.
Depending on the pretokenizer settings, a pretoken often cannot cross a change of script,
while a word span always can, so one span may cover several pretokens.

Spans are identified by Unicode general category and script, with no regular expression involved. \Cref{app:spans} gives the exact rules; here we give a high-level summary of the different types of spans: 
\begin{itemize}[nosep,leftmargin=*]
    \item A \emph{word span}
is a maximal run of letters and attached marks from a fixed list of 20 scripts that separate words with spaces, where runs from two such scripts that touch
form a single word span. 
\item A \emph{digit span} is a run of characters in the number category.
\item A \emph{punctuation span} is a run in the punctuation and format categories, together
with symbols other than emoji.
\item Each run of whitespace characters forms a span of its own, but only a single space can be elided.
\end{itemize}
All remaining spans, including those in scripts that do not separate words with spaces, such as Han, and emoji, are never marked.

\subsection{Word Boundary Encoding}

\begin{table*}
% Generated by paper_utils/boundary/make_example_table.py. Do not edit.
% Requires booktabs and the paper's \mk, \shf, \cps, \tsp, \pretokens and \tokens macros.
% sentence: Ash caught 3 SolidGoldMagikarp. WOW!
\centering
\small
\begin{tabular}{@{}l l r@{}}
\toprule
Scheme & Pre-tokens & \# \\
\midrule
\plainscheme & \pretokens{Ash,{\tsp}caught,\tsp,3,{\tsp}SolidGoldMagikarp,.,{\tsp}WOW,!} & 8 \\[3pt]
\bnd{w} & \pretokens{\mk Ash\mk,\mk caught\mk,\tsp,3,\tsp,\mk SolidGoldMagikarp\mk,.,\tsp,\mk WOW\mk,!} & 10 \\[3pt]
\bnd{wp} & \pretokens{\mk Ash\mk,\mk caught\mk,\tsp,3,\tsp,\mk SolidGoldMagikarp\mk,.\mk,\mk WOW\mk,!} & 9 \\[3pt]
\bnd{wpd} & \pretokens{\mk Ash\mk,\mk caught\mk,\mk 3\mk,\mk SolidGoldMagikarp\mk,.\mk,\mk WOW\mk,!} & 7 \\[3pt]
\bnd{wpdcaps} & \pretokens{\shf\mk ash\mk,\mk caught\mk,\mk 3\mk,\mk SolidGoldMagikarp\mk,.\mk,\cps\mk wow\mk,!} & 7 \\
\bottomrule
\end{tabular}
\caption{Pretokens under each scheme, before vocab entries are learned. Under
\bnd{w}, the space between two marked word spans is removed, but the spaces
bordering \texttt{3} and the space after the period remain, raising the count
from 8 to 10.
\bnd{wp} marks the period on its right only, where a space was removed. \bnd{wpd} marks the digit span as well, and no space remains.
\bnd{wpdcaps} places \shf{} or \cps{} before the span's opening marker, so the
count is unchanged and \tokens{\mk ash\mk} is a pretoken that largely overlaps with
the lower-case form.
Mixed case is not restorable from a single
marker, so \texttt{SolidGoldMagikarp} is left as-is.}
\label{tab:example}
    
\end{table*}

Our boundary-encoding scheme is inspired by work on reverse-engineering Anthropic's Claude
tokenizer~\citep{land2026claude-tokenizer}, though it differs from the scheme described
there. 

We add one atomic `boundary marker' token to the set of atomic tokens, which we denote
\mk{} throughout. Conceptually, encoding proceeds in three steps:
\begin{enumerate}[nosep,leftmargin=*,label=(\arabic*)]
\item Place markers on spans according to their type (rules described below).
\item Remove every single-space span that lies between two markers.
\item Pretokenize each span, with the markers included in its first and last pretoken.
\end{enumerate}
In practice we use a specialized pretokenizer that combines these steps, described in
\Cref{app:spans}.

Decoding reverses steps 2 and 1: replace every pair of adjacent markers with a
space, then remove every remaining marker. Decoding reconstructs the original
text whenever every pair of adjacent markers was produced by step~2. This is guaranteed by the following two properties. 

\paragraph{Word spans are marked on both sides.}
This requires that two word spans are never adjacent in the input without a space between them, which
holds by the definition of a word span: a run of letters that continues across a script change stays within one span. 
A span crossing a script change carries markers at its outer edges only, while the
pretokenizer can still split it at the script change: \dx{} is one span and two pretokens,
\tokens{\mk\ensuremath{\Delta},\texttt{x}\mk}. Were it two spans, it would be marked \mk$\Delta$\mk\mk x\mk{} and would decode to
\texttt{$\Delta$\tsp{}x}.
A word span with no space before it receives a marker with no marker adjacent
to it, which decoding removes: \texttt{the} in \texttt{"the} is marked
\texttt{"}\mk the\mk, giving the same pretoken \tokens{\mk the\mk} as would be given to the corresponding span in \texttt{the cat}. This is the duplication the scheme removes.

\begin{table}[t]
\small
\centering
\begin{tabular}{lll}
\toprule
Input & Encoding & Markers \\
\midrule
\texttt{the cat}   & \tokens{\mk the\mk,\mk cat\mk}        & --- \\
\texttt{the, cat}  & \tokens{\mk the\mk,{,\mk},\mk cat\mk}    & right \\
\texttt{the ,cat}  & \tokens{\mk the\mk,\mk{,},\mk cat\mk}    & left \\
\texttt{the , cat} & \tokens{\mk the\mk,\mk{,}\mk,\mk cat\mk} & both \\
\texttt{"the}      & \tokens{",\mk the\mk}                 & --- \\
\bottomrule
\end{tabular}
\caption{Marking of a punctuation span under \bnd{wp} and \bnd{wpd}.}
\label{tab:adjacency}
\end{table}

\paragraph{Punctuation and digit spans are marked only on a side bordering a
space.} These spans can be adjacent in the input with no space between them,
so marking both sides would produce pairs of adjacent markers that no removed
space accounts for. Under such marking, \texttt{the, cat} would be encoded as
\tokens{\mk the\mk,{\mk{,}\mk},\mk cat\mk}, which decodes to \texttt{the , cat} rather
than \texttt{the, cat}. 
\Cref{tab:adjacency} shows the cases explicitly.

We consider three schemes, which differ only in which categories of spans are marked: \bnd{w}
marks word spans, \bnd{wp} adds punctuation spans, and \bnd{wpd} adds digit spans. 
Each scheme introduces the possibility of duplicate entries similar to those under the leading-whitespace convention.
Under all three, a subword can occupy up to four entries, according to whether it carries a
marker on its left, its right, both, or neither, since the marker is attached to the first
piece and the last piece of a span.
This is arguably a desirable property, since subwords that occupy different positions within a word often encode different meanings, e.g., \texttt{ing} at the end of a word vs.\ in its middle.  
Nonetheless, this is a limitation of the method: while whole-word entries stop being duplicated, subwords still can be. 
We see a similar situation under \bnd{wp}: there can be entries for common punctuation both with and without a marker, and \bnd{wpd} adds the same for digits. 
Marking digit spans would also
interact with the practice of splitting digit runs into groups of three: a
digit run is one span, marked only at its outer edges, so the leading group
carries a marker and the interior groups do not, giving \tokens{\mk{}123} and
\tokens{123} as separate entries.\footnote{The tokenizers evaluated here do not group digits, nor do we evaluate on any math tasks, so the effects of this property are not yet known.}\looseness=-1

\subsection{Caps-code encoding}
\label{sec:caps}

The caps-code scheme also follows reverse-engineering work, along with tokenization 
examples published by Anthropic in~\citet{lindsey2025biology}. We add two more atomic tokens to the alphabet: \shf{} for
title case and \cps{} for upper case. 
In the same processing step where boundary markers are placed, a title case or upper case marker is attached before the opening marker of spans entirely in title case or upper case, respectively. All the characters in that span are then lowercased.  \texttt{Ash} is encoded as
\tokens{\shf\mk ash\mk} and \texttt{ASH} as \tokens{\cps\mk ash\mk}, so
\texttt{ash}, \texttt{Ash} and \texttt{ASH} can share the entry
\tokens{\mk ash\mk}. Note that caps codes are ignored in the space-removal step (2) of encoding: if only a single space and the caps code exist between two markers, that space is still removed. Likewise, when decoding, \tokens{\mk\shf\mk} and \tokens{\mk\cps\mk} decode to a space.

A code is used only when the span's case pattern is representable and the transformation is invertible: (i) the span is entirely in title case or entirely in upper case; (ii) lower-casing the span and then applying that case pattern returns the original span. 
\texttt{SolidGoldMagikarp} fails the first condition, so it is encoded as
\tokens{\mk SolidGoldMagikarp\mk}, as shown in \Cref{tab:example}. Condition (ii) can fail even when (i) holds, because lower-casing is not invertible for every character. The Turkish
\texttt{İ} is the only common example. It is entirely in title case, but its lower-case form does not map back. 
We use \shf{} for one-letter spans, which qualify for either code.

The caps codes likewise do not make duplicated entries impossible. A code and the first
pretoken of the span it applies to are not separated by the pretokenizer, so the training procedure can merge them and
produce \tokens{\shf\mk the\mk} as an entry distinct from \tokens{\mk the\mk}.
\Cref{tab:example} shows all tested schemes on an example sentence.

\section{Experimental Setup}

\begin{table*}[t]
% Generated by paper_utils/boundary/make_intrinsic_table.py. Do not edit.
% Requires booktabs, xcolor and the paper's \bnds and \plainscheme macros.
% source: paper_utils/boundary/paper/generated/eval_goldfish.json (quick corpus)
\centering
\small
\begin{tabular}{l rrrr rrrr}
\toprule
Scheme & \multicolumn{4}{c}{BPE} & \multicolumn{4}{c}{MinGram} \\
\cmidrule(lr){2-5} \cmidrule(lr){6-9}
 & \multicolumn{2}{c}{Compression} & \multicolumn{2}{c}{MorphScore} & \multicolumn{2}{c}{Compression} & \multicolumn{2}{c}{MorphScore} \\
\cmidrule(lr){2-3} \cmidrule(lr){4-5} \cmidrule(lr){6-7} \cmidrule(lr){8-9}
 & train & eval & credit & exclude & train & eval & credit & exclude \\
\midrule
%\plainscheme & \underline{4.04} & \textbf{4.10} & 0.19\,\textcolor{black!55}{(0.70)} & \textbf{0.11}\,\textcolor{black!55}{(0.13)} & \textbf{4.11} & \textbf{4.17} & 0.32\,\textcolor{black!55}{(0.81)} & \textbf{0.25}\,\textcolor{black!55}{(0.38)} \\
%\bnds{w} & $-8.70$ & $-9.30$ & \underline{0.657} & \underline{0.05} & $-9.19$ & $-9.85$ & \underline{0.691} & \underline{0.06} \\
%\bnds{wp} & $-1.82$ & $-2.19$ & 0.655 & \underline{0.05} & $-2.26$ & $-2.70$ & 0.690 & \underline{0.06} \\
%\bnds{wpd} & $-0.09$ & $-0.47$ & 0.654 & \underline{0.05} & $-0.52$ & $-0.96$ & 0.689 & \underline{0.06} \\
%\bnds{wpdcaps} & $\boldsymbol{+0.06}$ & $\underline{-0.33}$ & \textbf{0.659} & \underline{0.05} & $\underline{-0.07}$ & $\underline{-0.52}$ & \boldsymbol{$0.709$} & \underline{0.06} \\

\plainscheme & \underline{4.04} & \textbf{4.10} & 0.19\,\textcolor{black!55}{(0.70)} & \textbf{0.11}\,\textcolor{black!55}{(0.13)} & \textbf{4.11} & \textbf{4.17} & 0.32\,\textcolor{black!55}{(0.81)} & \textbf{0.25}\,\textcolor{black!55}{(0.38)} \\
\bnds{w} & $-$8.70 & $-$9.30 & \underline{0.657} & \underline{0.05} & $-$9.19 & $-$9.85 & \underline{0.691} & \underline{0.06} \\
\bnds{wp} & $-$1.82 & $-$2.19 & 0.655 & \underline{0.05} & $-$2.26 & $-$2.70 & 0.690 & \underline{0.06} \\
\bnds{wpd} & $-$0.09 & $-$0.47 & 0.654 & \underline{0.05} & $-$0.52 & $-$0.96 & 0.689 & \underline{0.06} \\
\bnds{wpdcaps} & \textbf{+0.06} & \underline{$-$0.33} & \textbf{0.659} & \underline{0.05} & \underline{$-$0.07} & \underline{$-$0.52} & \textbf{0.709} & \underline{0.06} \\

\bottomrule
\end{tabular}
\caption{Intrinsic evaluation results. \plainscheme{} compression is characters per token averaged over the six languages, every other compression cell the percentage change against it, higher is better. MorphScore is over English, higher is better, under both of its settings for words the tokenizer leaves whole: \emph{credit} scores them as correct, \emph{exclude} drops them and scores only the words that were split. For \plainscheme{}, both are also shown in grey with the gold word segmented with a leading space, which significantly affects scores. \textbf{Bold} is best in a column and \underline{underline} runner-up, grey figures excluded, counting \plainscheme{} as zero in the compression columns. Per-language numbers in \Cref{app:allthetokenizers}.}
\label{tab:intrinsic-main}

\end{table*}
\begin{table*}[t]

\centering
\small
\begin{tabular}{lrrrrrr}
\toprule
Scheme & \multicolumn{2}{c}{English} & \multicolumn{2}{c}{Korean} & \multicolumn{2}{c}{Russian} \\
\cmidrule(lr){2-3} \cmidrule(lr){4-5} \cmidrule(lr){6-7}
 & BPE & MinGram & BPE & MinGram & BPE & MinGram \\
\midrule
\plainscheme
  & 0.8853 & 0.8837 & \textbf{0.8543} & \textbf{0.8477} & 0.5503 & 0.5492 \\
\bnds{w} ($\Delta$)
  & \textbf{$-$0.0085} & $-$0.0074 & +0.0003 & +0.0027 & $-$0.0071 & $-$0.0058 \\
\bnds{wpd} ($\Delta$)
  & $-$0.0053 & $-$0.0032 & +0.0003 & +0.0023 & $-$0.0068 & $-$0.0066 \\
\bnds{wpdcaps} ($\Delta$)
  & $-$0.0060 & \textbf{$-$0.0077} & +0.0003 & +0.0021 & \textbf{$-$0.0081} & \textbf{$-$0.0093} \\
\bottomrule
\end{tabular}
\caption{Downstream language modeling in English, Korean and Russian.
The \plainscheme{} row reports mean validation bits per byte (BPB);
other rows report mean within-seed differences ($\Delta$) from
\plainscheme{}, over three seeds. Negative differences indicate improvement.
All English and Russian improvements are significant at $p<0.01$
(two-sided paired $t$-test) and none of the Korean differences are
significant at $p<0.05$.
\textbf{Bold} indicates the lowest BPB within each column.
BPB values are not comparable across languages.}
\label{tab:downstream-multilang}

\end{table*}

All tokenizers build on SCRIPT encoding \citep{scriptbpe}, which segments text
into maximal runs of shared Unicode script and supercategory (\cref{app:spans}) and represents each
character as a pair of tokens identifying its script-category block and its index within that block.
Vocabulary entries are learned over these pretokens. SCRIPT also defines which scripts separate
words with spaces, and we use that subset to determine which spans receive
a boundary marker.

\paragraph{Schemes compared.} We compare five schemes, which share the base encoding and
pretokenization, differing only in which spans receive boundary markers: (i) \plainscheme{}: the leading-space convention (baseline); 
(ii--iv) \bnd{w}, \bnd{wp}, and \bnd{wpd}: as described in \cref{sec:method}; (v) \bnd{wpdcaps}: \bnd{wpd} with the case codes of \Cref{sec:caps}.

%A word pretoken is delimited only at its outer edges, so \texttt{s$\pi$} keeps its pretokenization split and receives no internal marker, which would otherwise decode as a space.

\paragraph{Tokenizer training.}
We train each scheme with two vocabulary-learning algorithms: BPE \citep{sennrich-etal-2016-neural} and MinGram \citep{land2026mingram,land2025piecesdoesunigramtokenization}, a
minimum-token-count Unigram trainer initialized from BPE. 
Using both shows whether the results are specific to a tokenizer learning algorithm.
Every tokenizer has a total vocabulary size of 34{,}685 tokens, regardless of the boundary and case tokens, which differ by at most three entries between schemes.

\paragraph{Data and metrics.} 
Tokenizers are trained on 5\,GB of FineWeb/FineWeb-2 per language and evaluated on the monolingual
sets of \citet{chang2026goldfishmonolinguallanguagemodels} for English, German, Finnish, Russian, Arabic and
Korean, covering the Latin, Cyrillic, Arabic and Hangul scripts. We report
characters per token on both the training and held-out corpora. 
We also train tokenizers on English and Korean text together, using prefixes of the same
5\,GB samples, with mixture proportions measured in characters. 

\section{Results}\label{sec:results}

\paragraph{Intrinsic evaluation.} 

\Cref{tab:intrinsic-main} shows results for compression and MorphScore~\citep{arnett2025alignment,arnett-bergen-2025-language}.
The words-only boundary scheme \bnd{w} has substantially worse compression, but adding markers to punctuation spans closes most of the gap, and adding markers to digit spans further reduces it.
Introducing case codes is roughly neutral.
Under \plainscheme{}, space duplicates take 18--39\% of the vocabulary.
\Cref{app:allthetokenizers} gives per-language compression and duplication.

MorphScore measures agreement between predicted and gold
morpheme boundaries, penalizing both missed boundaries and
extra splits. We report two settings for words left whole:
crediting them as correctly segmented or excluding them
from evaluation.
Crediting them, the marker schemes score far above \plainscheme{}.
Excluding them, the ordering reverses. This closely tracks the share of gold words emitted as a single
token, which is $\approx65\%$ for the marker schemes and 8\% for \plainscheme{}.
Additionally, we find that the standard way to measure morphological alignment is highly affected by whether the gold word is segmented with or without a leading space, and we report both for \plainscheme{}, though for clarity we do not include the nonstandard space-prefixed measure in the rankings.

\paragraph{Downstream language modeling.}
To measure downstream performance, we train depth-12 nanochat
models with default hyperparameter settings. Models have a total of 298M parameters, with 6 attention heads, embedding dimension 768, and a context length of 2048 tokens.
English models are trained  on ClimbMix \citep{diao2025nemotronclimbclusteringbasediterativedata},
 using the English tokenizers. Korean and Russian models are trained with the Korean and Russian tokenizers, respectively, on a
sample from a filtered version of FineWeb-2 \citep{penedo2025fineweb2pipelinescale}, using the filtering protocol of Apertus~1.0 \citep{apertus2025}. We train three seeds per scheme and algorithm. Within a language, the training files are presented in the same order to all models with the same seed. 
We measure performance\footnote{We also measure DCLM CORE for English models, but results are too noisy at this model scale to support reliable comparisons.
Moreover, some boundary schemes violate the prefix-consistency assumption made by several CORE tasks.\looseness=-1} as bits per byte (BPB) using the original text length under UTF-8 encoding. 
In English and Russian, every evaluated marker scheme achieves significantly
lower bits per byte than \plainscheme{}
(\Cref{tab:downstream-multilang}; $p<0.01$, two-sided paired $t$-test
over three seeds).
The ordering is not the compression ordering: under \bnd{w},
English characters per token is 9\% lower than under \plainscheme{},
so a fixed token budget covers less text, yet \bnd{w} gives the
lowest bits per byte of the English BPE models.
In Korean, no marker scheme improves mean BPB over
\plainscheme{}, although none of the differences are statistically significant.

\vspace{-3pt}
\paragraph{Shared English and Korean vocabularies.}
To investigate whether reducing duplication frees vocabulary
capacity for other languages, we train tokenizers on English
and Korean together (\Cref{tab:bilingual-main,tab:bilingual-learned} in \cref{app:bilingual}).
Under \bnd{w}, characters per token is 8.6 to 9.8\%
lower than under \plainscheme{} in both languages, as in the monolingual tokenizers.
In the MinGram experiment with 64k learned entries, trained on 80\% English and 20\% Korean, \bnd{wpdcaps} shifts approximately
1{,}000 vocabulary entries from Latin toward Hangul,
demonstrating modest reallocation toward the
underrepresented language.

\section{Conclusion}
\label{sec:conclusion}

Our proposed boundary marker removes leading-space duplication and 
reduces duplication from capitalization, giving words a more 
canonical form across preceding contexts and common case patterns.
The markers introduce duplication of their own, since a subword at the start,
middle and end of a word gives separate entries, and the schemes with punctuation
marked end up closely matching the leading-space convention in terms of compression.

The benefit is elsewhere. A duplicated word has its occurrences divided between embeddings
that are trained independently, so the rarer form of a common word can end up as
under-trained as a genuinely rare string \citep{land-bartolo-2024-fishing}. In English
and Russian, bits per byte is lower than under \plainscheme{} for every marker scheme tested
downstream; in Korean it is not. We therefore recommend explicit boundary markers for both
words and punctuation where a canonical word form is wanted, with a test in each target
language, since we find no reduction in bits per byte in Korean.

\section*{Limitations}
\label{sec:limitations}

The language-modeling evaluation covers English, Korean and Russian at one model scale,
with three seeds per scheme and algorithm, and the English models are trained on a different
corpus from the Korean and Russian ones. Whether the markers lower bits per byte at larger
scale, or in German, Finnish and Arabic, is untested, and why Korean differs from English and
Russian is an open question. As \citet{bostrom-durrett-2020-byte} note, better compression
does not always produce a better language model.

Our experiments do not fully disentangle the effects of
reducing vocabulary duplication from those of explicitly
encoding word boundaries. For whole-word entries, removing
duplication allows more occurrences to update the same
embedding. For subwords, boundary markers also encode
position within a word, which may help distinguish
different uses of the same string, but may fragment the
training signal across positional variants.
The relative contributions of these mechanisms remain unclear.

Digit marking is the least settled part of the design. Delimiting whole digit runs gives every
number up to four marked forms. Splitting numbers into individual digits limits these to four
entries per \emph{digit} instead, 40 vocabulary entries in total, but lowers compression under both the plain and the marker schemes. 
No digit-splitting variant was tested in this work.

%\section*{Acknowledgments}

\bibliography{custom}

\clearpage
\onecolumn
\appendix
\crefalias{section}{appendix}

\section{Per-language tokenizer metrics}
\label{app:allthetokenizers}

\Cref{tab:intrinsic-quick} gives compression per language, and \Cref{tab:vocab-duplicates} shows vocabulary duplication statistics.

\begin{table*}[h!]
% Generated by paper_utils/boundary/make_intrinsic_table.py. Do not edit.
% Requires booktabs, xcolor and the paper's \bnds and \plainscheme macros.
% source: paper_utils/boundary/paper/generated/eval_goldfish.json (quick corpus)
\centering
\tiny
\setlength{\tabcolsep}{3pt}
\begin{tabular}{l rr rr rr rr rr rr rr}
\toprule

Scheme & \multicolumn{2}{c}{English} & \multicolumn{2}{c}{German} & \multicolumn{2}{c}{Finnish} & \multicolumn{2}{c}{Russian} & \multicolumn{2}{c}{Arabic} & \multicolumn{2}{c}{Korean} & \multicolumn{2}{c}{Mean} \\
\cmidrule(lr){2-3} \cmidrule(lr){4-5} \cmidrule(lr){6-7} \cmidrule(lr){8-9} \cmidrule(lr){10-11} \cmidrule(lr){12-13} \cmidrule(lr){14-15}
 & train & eval & train & eval & train & eval & train & eval & train & eval & train & eval & train & eval \\
\midrule

\multicolumn{15}{l}{\emph{BPE}} \\
\plainscheme & 4.334 & 4.456 & \textbf{4.541} & \textbf{4.573} & \textbf{4.646} & \textbf{5.010} & 4.336 & 4.316 & 4.098 & \textbf{3.968} & \textbf{2.285} & \textbf{2.296} & \underline{4.040} & \textbf{4.103} \\
\bnds{w} & $-$8.30 & $-$8.85 & $-$9.73 & $-$10.70 & $-$7.81 & $-$7.74 & $-$10.01 & $-$10.89 & $-$8.15 & $-$8.50 & $-$8.22 & $-$9.10 & $-$8.70 & $-$9.30 \\
\bnds{wp} & $-$0.95 & $-$1.29 & $-$2.50 & $-$3.00 & $-$1.72 & $-$1.67 & $-$0.97 & $-$1.27 & $-$1.30 & $-$1.92 & $-$3.48 & $-$3.98 & $-$1.82 & $-$2.19 \\
\bnds{wpd} & \underline{+0.82} & \underline{+0.63} & $-$0.67 & $-$1.07 & $-$0.27 & \underline{$-$0.48} & \underline{+0.79} & \underline{+0.48} & \underline{+0.42} & $-$0.23 & $-$1.67 & \underline{$-$2.15} & $-$0.09 & $-$0.47 \\
\bnds{wpdcaps} & \textbf{+1.02} & \textbf{+0.76} & \underline{$-$0.43} & \underline{$-$0.83} & \underline{$-$0.19} & $-$0.54 & \textbf{+1.14} & \textbf{+0.96} & \textbf{+0.46} & \underline{$-$0.19} & \underline{$-$1.64} & \underline{$-$2.15} & \textbf{+0.06} & \underline{$-$0.33} \\
\midrule
\multicolumn{15}{l}{\emph{MinGram}} \\
\plainscheme & 4.373 & 4.499 & \textbf{4.633} & \textbf{4.663} & \textbf{4.796} & \textbf{5.153} & 4.385 & 4.366 & \textbf{4.147} & \textbf{4.020} & \textbf{2.297} & \textbf{2.308} & \textbf{4.105} & \textbf{4.168} \\
\bnds{w} & $-$8.56 & $-$9.17 & $-$10.05 & $-$10.99 & $-$9.21 & $-$9.53 & $-$10.15 & $-$11.04 & $-$8.71 & $-$9.08 & $-$8.43 & $-$9.28 & $-$9.19 & $-$9.85 \\
\bnds{wp} & $-$1.19 & $-$1.57 & $-$2.72 & $-$3.18 & $-$3.10 & $-$3.50 & $-$1.03 & $-$1.32 & $-$1.86 & $-$2.51 & $-$3.68 & $-$4.14 & $-$2.26 & $-$2.70 \\
\bnds{wpd} & \underline{+0.61} & \underline{+0.38} & $-$0.84 & $-$1.19 & $-$1.64 & $-$2.32 & \underline{+0.76} & \underline{+0.46} & $-$0.13 & $-$0.82 & $-$1.86 & $-$2.30 & $-$0.52 & $-$0.96 \\
\bnds{wpdcaps} & \textbf{+1.06} & \textbf{+0.77} & \underline{$-$0.44} & \underline{$-$0.79} & \underline{$-$0.71} & \underline{$-$1.40} & \textbf{+1.56} & \textbf{+1.34} & \underline{$-$0.08} & \underline{$-$0.77} & \underline{$-$1.79} & \underline{$-$2.27} & \underline{$-$0.07} & \underline{$-$0.52} \\

\bottomrule
\end{tabular}
\caption{Compression, trained on 5\,GB of FineWeb per language, 32k matched learned tokens, evaluated on held-out Goldfish. \plainscheme{} is absolute characters per token on each corpus, every other cell the percentage change against it within the same block, higher is better. \textbf{Bold} is best in a column and \underline{underline} runner-up, counting \plainscheme{} as zero in the compression columns.}
\label{tab:intrinsic-quick}

\end{table*}

\begin{table*}[h!]
% Generated by paper_utils/boundary/vocab_duplicates.py. Do not edit.
% Requires booktabs, xcolor and the paper's \bnds and \plainscheme macros.
% source: paper_utils/boundary/paper/generated/vocab_duplicates.json (quick corpus)
\centering
\tiny
\setlength{\tabcolsep}{3pt}
\begin{tabular}{l rr rr rr rr rr rr rr}
\toprule
Scheme & \multicolumn{2}{c}{English} & \multicolumn{2}{c}{German} & \multicolumn{2}{c}{Finnish} & \multicolumn{2}{c}{Russian} & \multicolumn{2}{c}{Arabic} & \multicolumn{2}{c}{Korean} & \multicolumn{2}{c}{Mean} \\
\cmidrule(lr){2-3} \cmidrule(lr){4-5} \cmidrule(lr){6-7} \cmidrule(lr){8-9} \cmidrule(lr){10-11} \cmidrule(lr){12-13} \cmidrule(lr){14-15}
 & space & case & space & case & space & case & space & case & space & case & space & case & space & case \\
\midrule
\multicolumn{15}{l}{\emph{BPE}} \\
\plainscheme & 25.3 & 37.4 & 26.6 & 30.3 & 26.6 & 23.9 & 19.3 & 17.4 & 20.4 & 2.1 & 38.8 & 3.5 & 26.1 & 19.1 \\
\bnds{w} & 0.0 & 36.4 & 0.0 & 25.2 & 0.0 & 25.5 & 0.0 & 22.5 & 0.0 & 2.2 & 0.0 & 3.6 & 0.0 & 19.2 \\
\bnds{wp} & 0.0 & 36.2 & 0.0 & 25.1 & 0.0 & 25.5 & 0.0 & 22.4 & 0.0 & 2.2 & 0.0 & 3.6 & 0.0 & 19.2 \\
\bnds{wpd} & 0.0 & 35.8 & 0.0 & 24.8 & 0.0 & 25.2 & 0.0 & 22.1 & 0.0 & 2.2 & 0.0 & 3.5 & 0.0 & 19.0 \\
\bnds{wpdcaps} & 0.0 & 33.6 & 0.0 & 23.2 & 0.0 & 23.2 & 0.0 & 20.8 & 0.0 & 1.8 & 0.0 & 2.8 & 0.0 & 17.6 \\
\midrule
\multicolumn{15}{l}{\emph{MinGram}} \\
\plainscheme & 26.5 & 39.6 & 27.4 & 31.1 & 27.4 & 25.3 & 18.4 & 17.5 & 20.8 & 2.2 & 38.6 & 3.5 & 26.5 & 19.9 \\
\bnds{w} & 0.0 & 38.1 & 0.0 & 25.5 & 0.0 & 26.1 & 0.0 & 22.4 & 0.0 & 2.3 & 0.0 & 3.6 & 0.0 & 19.7 \\
\bnds{wp} & 0.0 & 37.8 & 0.0 & 25.4 & 0.0 & 26.0 & 0.0 & 22.4 & 0.0 & 2.4 & 0.0 & 3.5 & 0.0 & 19.6 \\
\bnds{wpd} & 0.0 & 37.5 & 0.0 & 25.2 & 0.0 & 25.8 & 0.0 & 22.1 & 0.0 & 2.3 & 0.0 & 3.5 & 0.0 & 19.4 \\
\bnds{wpdcaps} & 0.0 & 33.7 & 0.0 & 22.9 & 0.0 & 22.3 & 0.0 & 17.8 & 0.0 & 1.9 & 0.0 & 2.7 & 0.0 & 16.9 \\
\bottomrule
\end{tabular}
\caption{Vocabulary entries duplicating another entry, as a percentage of
vocabulary size. \emph{space}: differing only by a leading space.
\emph{case}: differing only by capitalization, including a case code plus
its span. Both entries of a pair count, and the measures overlap.
Marker schemes are zero for \emph{space} by construction.
Case codes lower \emph{case} without clearing it, as mixed case is left
literal and frequent title-case forms still earn entries.}
\label{tab:vocab-duplicates}

\end{table*}

\clearpage

\section{Shared English and Korean vocabularies}
\label{app:bilingual}

\Cref{tab:bilingual-main} gives the BPE and MinGram tokenizers trained on 2B characters with
50\% English, and \Cref{tab:bilingual-learned} gives the MinGram tokenizers with 64{,}000
learned entries, trained on 5B characters with 80\% English.

\begin{table*}[h!]
% Generated by paper_utils/boundary/make_bilingual_tables.py. Do not edit.
% Requires booktabs and the paper's \bnds and \plainscheme macros.
% source: paper_utils/boundary/paper/generated/bilingual_tables.json, extract of bilingual_results.json
\centering
\small
\begin{tabular}{lrrrrrrrr}
\toprule
 & \multicolumn{2}{c}{Characters per token} & \multicolumn{4}{c}{Learned entries by script} & \multicolumn{2}{c}{Duplicates (\%)} \\
\cmidrule(lr){2-3} \cmidrule(lr){4-7} \cmidrule(lr){8-9}
Scheme & English & Korean & Latin & Hangul & Other & Nonlex. & space & case \\
\midrule
\multicolumn{9}{l}{\emph{BPE}} \\
\plainscheme & 4.069 & 2.193 & 12,443 & 19,673 & 116 & 743 & 34.9 & 12.9 \\
\bnds{w} & $-9.21$ & $-8.97$ & $+8$ & $+138$ & $-25$ & $-122$ & 0.0 & 12.7 \\
\bnds{wcaps} & $-9.17$ & $-8.90$ & $-109$ & $+246$ & $-22$ & $-118$ & 0.0 & 11.4 \\
\bnds{wp} & $-2.19$ & $-4.22$ & $-49$ & $+55$ & $-26$ & $+19$ & 0.0 & 12.6 \\
\bnds{wpcaps} & $-2.16$ & $-4.16$ & $-154$ & $+150$ & $-24$ & $+25$ & 0.0 & 11.4 \\
\bnds{wpd} & $-0.61$ & $-2.57$ & $-138$ & $-90$ & $-26$ & $+253$ & 0.0 & 12.5 \\
\bnds{wpdcaps} & $-0.56$ & $-2.50$ & $-243$ & $+5$ & $-24$ & $+259$ & 0.0 & 11.3 \\
\midrule
\multicolumn{9}{l}{\emph{MinGram}} \\
\plainscheme & 4.106 & 2.212 & 11,857 & 20,235 & 117 & 766 & 35.3 & 13.0 \\
\bnds{w} & $-9.83$ & $-9.00$ & $-119$ & $+252$ & $-18$ & $-116$ & 0.0 & 12.2 \\
\bnds{wcaps} & $-9.12$ & $-8.88$ & $-344$ & $+459$ & $-9$ & $-109$ & 0.0 & 10.2 \\
\bnds{wp} & $-2.86$ & $-4.23$ & $-166$ & $+142$ & $-18$ & $+41$ & 0.0 & 12.2 \\
\bnds{wpcaps} & $-2.05$ & $-4.09$ & $-385$ & $+344$ & $-13$ & $+51$ & 0.0 & 10.2 \\
\bnds{wpd} & $-1.24$ & $-2.57$ & $-238$ & $-7$ & $-18$ & $+262$ & 0.0 & 12.1 \\
\bnds{wpdcaps} & $-0.43$ & $-2.43$ & $-444$ & $+187$ & $-13$ & $+267$ & 0.0 & 10.1 \\
\bottomrule
\end{tabular}
\caption{Tokenizers trained on English and Korean together, total vocabulary 34,685, on 2B characters, 50\% English, with the English share measured in characters.
Characters per token on 500,000 characters of English and 499,991 of
Korean Goldfish text: absolute in the \plainscheme{} row, percentage change against
it in the other rows, positive meaning fewer tokens.
Learned entries by script: entries beyond the atomic alphabet,
grouped by the script of their letters, with \emph{Other} for letters of any other
script and \emph{Nonlex.} for entries without letters; counts in the \plainscheme{}
row, change in count in the other rows.
Duplicates: learned entries that differ from another learned entry only by a
leading space (\emph{space}) or only by capitalization (\emph{case}), as a
percentage of learned entries, both entries of a pair counted. Unlike
\Cref{tab:vocab-duplicates}, atomic tokens are excluded, and \emph{space}
includes whitespace-only pairs such as a newline and a space followed by a newline.}
\label{tab:bilingual-main}

\end{table*}

\begin{table*}[h!]
% Generated by paper_utils/boundary/make_bilingual_tables.py. Do not edit.
% Requires booktabs and the paper's \bnds and \plainscheme macros.
% source: paper_utils/boundary/paper/generated/bilingual_tables.json, extract of bilingual_results_learned64000.json
\centering
\small
\begin{tabular}{lrrrrrrrr}
\toprule
 & \multicolumn{2}{c}{Characters per token} & \multicolumn{4}{c}{Learned entries by script} & \multicolumn{2}{c}{Duplicates (\%)} \\
\cmidrule(lr){2-3} \cmidrule(lr){4-7} \cmidrule(lr){8-9}
Scheme & English & Korean & Latin & Hangul & Other & Nonlex. & space & case \\
\midrule
\multicolumn{9}{l}{\emph{2B characters, 50\% English}} \\
\plainscheme & 4.368 & 2.423 & 21,530 & 40,557 & 420 & 1,493 & 37.0 & 14.1 \\
\bnds{w} & $-9.56$ & $-8.69$ & $-384$ & $+606$ & $-42$ & $-180$ & 0.0 & 13.1 \\
\bnds{wcaps} & $-9.16$ & $-8.64$ & $-782$ & $+978$ & $-33$ & $-163$ & 0.0 & 11.1 \\
\bnds{wp} & $-2.04$ & $-3.34$ & $-454$ & $+463$ & $-43$ & $+34$ & 0.0 & 13.1 \\
\bnds{wpcaps} & $-1.56$ & $-3.27$ & $-839$ & $+831$ & $-34$ & $+42$ & 0.0 & 11.1 \\
\bnds{wpd} & $-0.21$ & $-1.44$ & $-508$ & $+306$ & $-44$ & $+246$ & 0.0 & 13.0 \\
\bnds{wpdcaps} & $+0.28$ & $-1.35$ & $-903$ & $+677$ & $-39$ & $+265$ & 0.0 & 11.1 \\
\midrule
\multicolumn{9}{l}{\emph{5B characters, 80\% English}} \\
\plainscheme & 4.535 & 2.321 & 34,065 & 28,028 & 318 & 1,589 & 34.9 & 23.6 \\
\bnds{w} & $-9.48$ & $-8.65$ & $-671$ & $+852$ & $-20$ & $-161$ & 0.0 & 22.1 \\
\bnds{wpd} & $+0.34$ & $-1.73$ & $-894$ & $+594$ & $-26$ & $+326$ & 0.0 & 22.0 \\
\bnds{wpdcaps} & $+0.66$ & $-1.58$ & $-1{,}374$ & $+1{,}034$ & $-13$ & $+353$ & 0.0 & 19.4 \\
\bottomrule
\end{tabular}
\caption{MinGram tokenizers trained on English and Korean together, with 64,000 learned entries each, on the two mixtures named in the block headings, with English shares measured in characters.
Characters per token on 500,000 characters of English and 499,991 of
Korean Goldfish text: absolute in the \plainscheme{} row, percentage change against
it in the other rows, positive meaning fewer tokens.
Learned entries by script: entries beyond the atomic alphabet,
grouped by the script of their letters, with \emph{Other} for letters of any other
script and \emph{Nonlex.} for entries without letters; counts in the \plainscheme{}
row, change in count in the other rows.
Duplicates: learned entries that differ from another learned entry only by a
leading space (\emph{space}) or only by capitalization (\emph{case}), as a
percentage of learned entries, both entries of a pair counted. Unlike
\Cref{tab:vocab-duplicates}, atomic tokens are excluded, and \emph{space}
includes whitespace-only pairs such as a newline and a space followed by a newline.}
\label{tab:bilingual-learned}

\end{table*}

\clearpage
\section{Details of SCRIPT Tokenization and Boundary Marker Implementation}
\label{app:spans}

\subsection{SCRIPT v3}
We use the most recent version of SCRIPT (v3), which we briefly describe below.
Note that this version differs in some details from the version introduced in~\citet{scriptbpe}.

SCRIPT-based pretokenization represents each character by a pair of atomic tokens: a
\emph{block} token naming its script and folded category, and an \emph{index} token giving
its position within that block. 

The block token is based on Unicode properties.
Unicode gives every character a \emph{general category}, a
two-letter code saying what kind of character it is: \texttt{L} for letters, \texttt{M}
for marks that attach to a preceding letter such as accents and vowel signs, \texttt{N}
for numbers, \texttt{P} for punctuation, \texttt{S} for symbols, \texttt{Z} for separators
including the space, and \texttt{C} for controls and format characters. It also gives
every character a script, such as Latin, Cyrillic or Han.

SCRIPT v3 uses these properties to define larger classes of characters, which are then used
to encode text.
The Unicode general category is used to assign each character to one of the following \emph{supercategories}:
\begin{itemize}[nosep,leftmargin=*]
\item \texttt{LM} for letters and marks,
\item \texttt{N} for numbers,
\item \texttt{ZC} for whitespace, separators and controls,
\item \texttt{So} for Unicode's symbol-other category, covering emoji as well as e.g.\ \texttt{\textcopyright} and \texttt{\textdegree},
\item \texttt{PSF} for punctuation, the remaining symbols and format characters.
\end{itemize}

Characters in \texttt{LM} among 28 \emph{high-resource} scripts\footnote{The 20 space-using scripts Latin, Arabic, Devanagari, Hangul, Ethiopic,
Cyrillic, Greek, Hebrew, Bengali, Syriac, Oriya, Tamil, Telugu, Gurmukhi, Gujarati,
Sinhala, Malayalam, Armenian, Kannada and Georgian, plus Han, Hiragana, Katakana, Thai,
Myanmar, Khmer and Lao, and the special \emph{Common} script, which Unicode assigns to
characters shared between scripts, such as the digits \texttt{0}--\texttt{9}, ASCII
punctuation and the space itself.
Unicode records no property for whether a script is high-resource, or separates words with spaces, so these are fixed lists.}
are assigned a block based on both their script and their supercategory. Other
high-resource characters discard the script and are assigned to a supercategory block, so
Latin and Cyrillic letters are separate blocks while Arabic-Indic and Devanagari digits
share one block with \texttt{012}. Characters outside the high-resource set discard the
supercategory instead, so Tibetan letters, digits and punctuation are all one block. This
eliminates a number of very small blocks, reducing the number of SCRIPT tokens from
1{,}916 to 1{,}710.

\subsection{SCRIPT Pretokenization}

Pretokens are formed by grouping adjacent characters that share a script and supercategory, so no pretoken
spans a script change and no merge can cross one. Inherited pseudo-script characters join
the group before them.
A single space joins the group that follows it, if that
group's supercategory admits a leading space, specifically \texttt{LM} among the 20
space-using scripts and \texttt{PSF}. Nothing else does: a run of two or more spaces stands
alone, as does a space before e.g.\ emojis, digit runs, or Han, kana and Thai letters.

\begin{center}
\texttt{the sample weighs 42 $\mu$g?} \quad$\rightarrow$\quad
\tokens{the,\tsp sample,\tsp weighs,{\tsp},42,\tsp$\mu$,g,?}
\end{center}

\subsection{Boundary Marking}

Spans are formed from the same character groups, with adjacent word runs additionally
merged across script changes, and classified as:

\begin{itemize}[nosep,leftmargin=*]
\item \emph{word}: supercategory \texttt{LM} in a space-using script,
\item \emph{digit}: supercategory \texttt{N},
\item \emph{punctuation}: supercategory \texttt{PSF},
\item \emph{space}: a single space character,
\item \emph{other}: everything else.
\end{itemize}

Merging word runs across script changes is what makes the scheme well defined. Delimiting
each script run separately would put two markers together at the script change, which is
the same signal as an elided space, and decoding would insert a space that was not there.
After merging, two word spans are never adjacent.

\end{document}